\documentclass[12pt]{article}
\usepackage{amsmath}
\usepackage{graphicx}
\usepackage{enumerate}
\usepackage[sectionbib]{natbib}
\usepackage{url}
\newcommand{\blind}{1}

\usepackage{amsfonts,amssymb,amsthm}
\usepackage{epstopdf}
\usepackage[]{hyperref}
\usepackage[linesnumbered,ruled,vlined]{algorithm2e}
\usepackage{setspace}
\usepackage{caption,subcaption}
\usepackage{multirow}
\newtheorem{The}{Theorem}
\newtheorem{Pro}{Proposition}
\newtheorem{Cor}{Corollary}
\newtheorem{Exa}{Example}
\newtheorem{remark}{Remark}

\def \d {\mathbf{d}} \def \k {{\bf k}} \def \u {\mathbf{u}} \def \p {\mathbf{p}} \def \w {\mathbf{w}}
\def \one {{\bf 1}} \def \two {{\bf 2}} \def \three {{\bf 3}} \def \four {{\bf 4}} \def \five {{\bf 5}}
\def \six {{\bf 6}} \def \seven {{\bf 7}} \def \eight {{\bf 8}} \def \nine {{\bf 9}} \def \ten {{\bf 10}}
\def \var {\text{Var}}
\def \cov {\text{Cov}}
\def \E {\text{E}}
\def \x {\textbf{x}}

\def \Sh {\mathrm{Sh}}
\def \nc {\mathrm{nc}}
\usepackage{color}

\allowdisplaybreaks[4]
\begin{document}

\def\spacingset#1{\renewcommand{\baselinestretch}%
{#1}\small\normalsize} \spacingset{1}

\captionsetup[figure]{labelsep=period,singlelinecheck=off}
\captionsetup[table]{labelsep=period,singlelinecheck=off}


\if1\blind
{
  \title{\bf Fast Approximation of the Shapley Values Based on Order-of-Addition Experimental Designs}
  \author{Liuqing Yang$^1$, Yongdao Zhou$^1$, Haoda Fu$^2$, Min-Qian Liu$^1$, Wei Zheng$^3$
\\
$^1$ {\small School of Statistics and Data Science, LPMC $\&$ KLMDASR,} \\
{\small Nankai University, Tianjin 300071, China}\\
$^2$ {\small Global Statistical Sciences, Eli Lilly and Company,}\\ {\small Indianapolis, Indiana 46285, USA}\\
$^3$ {\small Haslam College of Business, University of Tennessee,}\\ {\small Knoxville, TN 37996, USA}\\
{\small Corresponding author Wei Zheng wzheng9@utk.edu}
}
  \date{}\maketitle
} \fi

\if0\blind
{
  \bigskip
  \bigskip
  \bigskip
  \begin{center}
    {\LARGE\bf Fast Approximation of the Shapley Values Based on Order-of-Addition Experimental Designs}
\end{center}
  \medskip
} \fi

\bigskip
\begin{abstract}
Shapley value is originally a concept in econometrics to fairly distribute both gains and costs to players in a coalition game. In the recent decades, its application has been extended to other areas such as marketing, engineering and machine learning. For example, it produces reasonable solutions for problems in sensitivity analysis, local model explanation towards the interpretable machine learning, node importance in social network, attribution models, etc. However, its heavy computational burden has been long recognized but rarely investigated. Specifically, in a $d$-player coalition game, calculating a Shapley value requires the evaluation of $d!$ or $2^d$ marginal contribution values, depending on whether we are taking the permutation or combination formulation of the Shapley value. Hence it becomes infeasible to calculate the Shapley value when $d$ is reasonably large. A common remedy is to take a random sample of the permutations to surrogate for the complete list of permutations. We find an advanced sampling scheme can be designed to yield much more accurate estimation of the Shapley value than the simple random sampling (SRS). Our sampling scheme is based on combinatorial structures in the field of design of experiments (DOE), particularly the order-of-addition experimental designs for the study of how the orderings of components would affect the output. We show that the obtained estimates are unbiased, and can sometimes deterministically recover the original Shapley value. Both theoretical and simulations results show that our DOE-based sampling scheme outperforms SRS in terms of estimation accuracy. Surprisingly, it is also slightly faster than SRS. Lastly, real data analysis is conducted for the C. elegans nervous system and the 9/11 terrorist network.
\end{abstract}

\noindent%
{\it Keywords:} Component orthogonal array; Game theory; Latin square; Local model explanation; Sensitivity analysis.
\vfill

\newpage
\spacingset{1.9} 
\section{Introduction}\label{intro}
The Shapley value, proposed by \cite{S1953}, is the unique method of allocating the group profit to each participant of a coalition that satisfies four axioms.
 Analogously to the profit allocation, the Shapley value can also be used in cost games to attribute costs among collaborators. See, e.g., \cite{SZ1996}, \cite{KS2005}, and \cite{ADKTWR2008}. In recent years, the application of the Shapley value has been further extended to the sensitivity analysis and local model explanation. 
Traditionally, sensitivity analysis is mainly carried out by methods based on the analysis of variance (ANOVA) decomposition, such as first-order effect, total effect, and Stone-Hooker ANOVA, see \cite{S1994}, \cite{H2007} and \cite{SRACCGST2008}. However, these methods have two shortcomings, neither of which the Shapley value possesses. First, they can only be defined under restrictive conditions for the input distributions, which even exclude the correlated Gaussian distribution. The other issue is the possibility of getting negative values for the importance measure.  
{To measure the importance of variables, \cite{LMG1980} assigned $R^2$ of the model to each variable using the Shapley value method, while \cite{O2014} and \cite{SNS2016} calculated the Shapley values according to variance.}
\cite{O2014} showed that the sum of the Shapley values equals the correct total variance if the inputs are independent. \cite{SNS2016} extended this conclusion to the general dependent case. \cite{OP2017} pointed out that the Shapley values are always nonnegative when the value function is defined as \cite{O2014} and \cite{SNS2016}. Local model explanation is an important topic for interpretable machine learning. It helps the understanding of how the black-box output is influenced by each individual input by computing variable importance values for each input of the model. For example, if a person is rejected by the loan decision model, the reason for his failure can be found through the local model explanation. \cite{LL2017} showed that the Shapley value is the only consistent and locally accurate variable importance value.

A well known challenge in implementing the Shapley value method is that its calculation becomes infeasible when the number of players, say $d$, is too large. \cite{DP1994} proved that computing the Shapley value is an NP-complete problem. There have been many efforts in literature to tackle this computational challenge. \cite{KHMR2004} estimate the Shapley value by predicting all the $2^d$ value functions based on partial calculated value functions. However, this method can become computationally infeasible when $d$ is large. \cite{FWJ2008} proposed an approximation method based on a particular property of value functions in voting games, which however cannot be applied to general situations. \cite{SCCP2014} proposed to divide players into clusters and aggregate the cluster-level exact Shapley values into an ultimate estimate of the original Shapley value. But this method relies on the assumption that the players can be well separated into subgroups, and it is not feasible when the number of players is large. \cite{LEL2018} proposed an algorithm to estimate the Shapley value in local model explanation, which can only work when the prediction model is a tree-ensemble model and the value function is the conditional expectation.  \cite{G2021} provided an efficient approximate algorithm by estimating the difference of two value functions directly, however, this method can only be used for sensitivity analysis especially with independent variables.

One approach for approximating the Shapley value in a general situation is to take a random sample of permutations of players, in view of the representation of the Shapley value as an average over all possible $d!$ permutations, as we show below. See the CGT and SNS algorithms proposed by \cite{CGT2009} and \cite{SNS2016}, respectively. Both algorithms employ simple random sampling (SRS). \cite{vHHL2018} suggested to make algorithmic adjustment on the permutations from the SRS, which is called the structured random sampling (StrRS) therein. The main idea of StrRS is to first obtain permutations by SRS and then swap the components within the permutations to achieve partial position balance. We show that this improvement was limited, even though the computation time is twice that of SRS.

We argue that there exist intrinsic relationships among the permutations so that it is desirable for a sample of permutations to exhibit certain space-filling structures and to well represent the whole permutation space. To this point, a seemingly unrelated field, design of experiment, serves as a library of possible good structures. We find that the order-of-addition (OofA) experimental designs are well connected to the problem of designing the sampling schemes of the permutations.

The rest of this paper is organized as follows. Section \ref{pre} provides some basic knowledge about the Shapley value and OofA experimental designs.
Section \ref{method} proposes the algorithm for estimating the Shapley value by using OofA experimental designs.
Theoretical properties of the obtained estimates are also derived.
Section \ref{sim} and Section \ref{realdata} illustrate our method and compare it with existing methods through simulations and real data analysis.
Section \ref{conc} concludes the paper with some remarks. All proofs and some additional illustrations are deferred to the Supplementary Material.

\section{Preliminaries}\label{pre}
\subsection{The Shapley Value} \label{pre.sh}
The Shapley value was proposed by \cite{S1953} to solve the income distribution problem in cooperative games. Let $\{1:d\}=\{1,2,\ldots,d\}$ be the set of $d$ players in a cooperative game, and $\u^c=\{1:d\}\backslash \u$ be the complementary set of $\u\subseteq \{1:d\}$. Denote by $\nu(\u)$ the value function, also called as characteristic function, of the player set $\u$ that measures the contribution generated by the set, where $\u=\{j_1,j_2,\ldots,j_{|\u|}\}$, and $|\u|$ is the size of the set $\u$. It is always assumed that $\nu(\emptyset)=0$. For a given permutation of $1,2,\ldots,d$, say $\pi$, the marginal contribution of player $j$ can be defined as $\Delta(\pi,\nu)_j=\nu(P_{\pi}^j\cup\{j\})-\nu(P_{\pi}^j)$, where $P_{\pi}^j$ is the set of predecessors of $j$ in the permutation $\pi$. The Shapley value of $j$ can be calculated as these marginal {contributions} averaged over $\Pi$, the set of all possible permutations, that is
\begin{equation}\label{eq.1}
\Sh_j(\nu)=\frac{1}{d!}\sum_{\pi\in\Pi}\Delta(\pi,\nu)_j,\quad j=1,2,\ldots,d.
\end{equation}
There is an alternative representation of the Shapley value
\begin{equation}\label{eqn:0109}
\Sh_j(\nu)= \frac{1}{d}\sum_{\u\subseteq\{j\}^c}{{d-1}\choose{|\u|}}^{-1}
\widetilde\Delta(\u,\nu)_j,\quad j=1,2,\ldots,d,
\end{equation}
where $\widetilde\Delta(\u,\nu)_j=\nu(\u\cup\{j\})-\nu(\u)$ is the marginal contribution of $j$ in the set $\u\cup\{j\}$. Without ambiguity, $\Sh_j(\nu)$ and $\Delta(\u,\nu)_j$ will be simply written as $\Sh_j$ and $\Delta(\u)_j$, respectively. See \cite{HM1989} for the equivalence of \eqref{eq.1} and \eqref{eqn:0109}, and \cite{R1988} for the calculation of the Shapley value. The calculations based on permutations in (\ref{eq.1}) allow us to use the OofA design structure for its approximation and hence this will be the focus of this paper.

\cite{S1953} proved that the Shapley value is the unique real function satisfying the following four axioms:\\
(1) Efficiency: $\sum_{j=1}^d \Sh_j=\nu(\{1:d\})$.\\
(2) Symmetry: If $\nu(\u\cup\{i\})=\nu(\u\cup\{j\})$ for all $\u\subseteq\{1:d\}\backslash\{i,j\}$, then $\Sh_i=\Sh_j$.\\
(3) Dummy: If $\nu(\u\cup\{j\})=\nu(\u)$ for all $\u\subseteq\{1:d\}$, then $\Sh_j=0$.\\
(4) Additivity: If $\nu(\cdot)$ and $\nu'(\cdot)$ have Shapley values $\Sh_j$ and $\Sh{'}_j$ respectively, then the game with value $\nu(\cdot)+\nu'(\cdot)$ has Shapley value $\Sh_j+\Sh{'}_j,\ j=1,2,\ldots,d$.\\
The efficiency axiom ensures that players distribute the whole resources available to the coalition. The symmetry axiom ensures players are fairly treated. The dummy axiom shows that players with no marginal contribution should not be rewarded. Lastly the additivity axiom requires that the value is an additive operator on the space of all games. More details of the Shapley value in game theory can be found in \cite{W2002}.

{Recently, the Shapley value is also applied in fields beyond game theory by choosing appropriate value functions. In this paper,
we emphasize two importance uses of these value functions, for sensitivity
analysis and for local model explanation, each of which we now
introduce.}

The primary objective of sensitivity analysis is to quantify the importance of inputs to a function. Suppose we have an input-output relationship as $y=g(\x)$, and let $F$ be the distribution of the input $\x$ with its $i$th component marginal distribution denoted by $F_i$. \cite{O2014} suggested doing sensitivity analysis with the Shapley value, and \cite{SNS2016} modified the value function as $\nu(\u)=\E[\var(Y|\x_{\u^c})]$, where $\x_{\u^c}=(x_{j_1},x_{j_2},\ldots,x_{j_{|\u^c|}})$, and $\u^c=\{1:d\}\backslash\u=\{j_1,j_2,\ldots,j_{|\u^c|}\}$ is the complementary set of  $\u\subset\{1:d\}$. The Shapley value with this value function is equal to that in \cite{O2014}, meanwhile estimations based on the former tend to be more accurate.

{The primary objective of local model
explanation is to explain any unit's output by its input values. For
example, we can find the characteristics that have an important impact on
the survival results of a specific passenger by analyzing the passenger
information on the Titanic. Such local model explanation has become one
of the important tools for interpretable machine learning. Both sensitivity
analysis and local model explanation can be achieved by the Shapley
value.} Several value functions are defined from different perspectives, which lead to different Shapley values, such as the conditional expectation Shapley \citep{SK2013}, the baseline Shapley \citep{SN2020} and the cohort Shapley \citep{MOS2020}. Users can choose one of them to use according to the analysis purpose and the sparsity of the data set.

Equation (\ref{eq.1}) implies that $\Sh_j$ is the population mean of all $d!$ marginal contributions of player $j$. An intuitive approach for its fast approximation is to estimate the population mean by a sample mean. Denote by $\pi_1,\ldots,\pi_m$ the $m$ permutations sampled from $\Pi$, one way to approximate $\Sh_j$ is by

\begin{equation}\label{eq.tildesh}
	\widehat{\Sh}^{SRS}_j=\frac{1}{m}\sum_{i=1}^m\Delta(\pi_i)_j.
\end{equation}
SRS is the most general and flexible sampling method. See the CGT and SNS a
lgorithms in \cite{CGT2009} and \cite{SNS2016} under the context of game theory and sensitive analysis, respectively. In fact, the CGT algorithm can also be adapted to estimate the conditional expectation Shapley, the baseline Shapley and the cohort Shapley. However, there has been study on alternative sampling scheme of the permutations other than SRS, see \cite{vHHL2018}. Here we propose to use the design structures in OofA experiments to improve on the sampling scheme.

\subsection{OofA Experimental Design} \label{pre.oofa}
When the response may be affected by the addition order of components, OofA experimental designs should be considered. Each run of an OofA experimental design is a permutation of $1,2,\ldots,d$.The design region consists of all the $d!$ possible permutations of $d$ components.
When $d$ is large, it is usually infeasible to run all $d!$ runs, so a fraction of the full OofA design is used instead.
The latter can be viewed as a designed selection of permutations $\pi$ in (\ref{eq.1}) for the estimation of the Shapley value.
To obtain efficient fractional OofA experimental designs, \cite{V2019} and \cite{PML2019} provided some construction methods based on the pair-wise ordering model proposed by \cite{V1995}. Their designs are {balanced} regarding relative ordering among the components, that is, the frequency with which component $i$ precedes component $j$ is a constant for any $i\neq j$. \cite{YSX2021} proposed a new type of OofA experimental designs, called component orthogonal arrays (COAs), which is {balanced} regarding the absolute ordering among pairs of the components, that is, the frequency of any pair of components at any two distinct positions is equivalent. Since our interest is on estimating average output across all the permutations instead of the relationship of how the relative positioning are affecting the output, COA could be more suitable for estimating the Shapley value.

Denote by COA$(r,d)$ an $r\times d$ array with entries from $\{1:d\}$, whose row is a permutation of $1,2,\ldots,d$, and for any two columns, each of the level combinations $(i,j)$ with $i\neq j$ and $i,j=1,2,\ldots,d$ appears $\lambda$ times, where the positive integer $\lambda=r/\left(d(d-1)\right)$ is the index. {When $d$ is a prime power, \cite{YSX2021} provided a construction method to obtain COA$(d(d-1),d)$ as follows. Let $GF(d)$ be a Galois field of order $d$, where $d=(d')^r$, $r$ is an integer and $d'$ is a prime. For $r\geq 1$, $GF(d)=\{a_0+a_1x+\ldots+a_{r-1}x^{r-1}:\ a_0,a_1,\ldots,a_{r-1}\in GF(d')\}=\{\alpha_0,\alpha_1,\ldots,\alpha_{d-1}\}$ with $\alpha_0=0$, and $GF(d')=\{0,1,\ldots,d'-1\}$. Let $M_0=(\alpha_0,\alpha_1,\ldots,\alpha_{d-1})'(\alpha_0,\alpha_1,\ldots,\alpha_{d-1})$ be a $d\times d$ matrix, where the multiplication is defined on $GF(d)$. Then generate $M$ by deleting the first row of $M_0$, and let
\begin{equation}\label{coa}
D=\begin{pmatrix}
M\oplus0\\ M\oplus1\\ \vdots\\ M\oplus d-1
\end{pmatrix},
\end{equation}
where $M\oplus i$ means that $i$ is added to each entry of $M$ and $\oplus$ in the parentheses is defined on $GF(d)$. A COA$(d(d-1),d)$ can be obtained by mapping the elements of $D$ to $\{1,2,\ldots,d\}$.} \cite{YSX2021} proved that, $(d-2)!$ different COA$(d(d-1),d)$ that share no common runs can be constructed by permuting the last $d-2$ columns of $D$ in equation \eqref{coa}, which can be regarded as $(d-2)!$ partitions of the full OofA design.

One restriction of the COA is that  $d$ is required to be a prime power. We shall propose two solutions when a COA does not exist. One is to augment the coalition with additional null players so that the new number of players is a prime power. See Section \ref{flexible} for the exploration of such approach, along with an example in Section \ref{realdata}.
The other solution is to use an alternative design. In fact, the Latin square (LS) as also used in OofA experiments always exist for any value of $d$. An LS of order $d$ is a $d\times d$ matrix in which each row and each column is a permutation of $1,2,\ldots,d$.
Since there are only $d$ permutations in an LS of order $d$, the LS is a good choice when $d$ is large or the computational complexity of the value function is high such that using a $COA(d(d-1),d)$ is unaffordable.
An LS is called the standard Latin square (SLS) if the first column of it is $(1,2,\ldots,d)'$. Note that there is always more than one SLS of order $d,\ d\geq4$. Denote by $S(d)$ the number of distinct SLSs of order $d$, then all $S(d)d!(d-1)!$ LSs of order $d$ can be obtained by carrying out all row permutations and column permutations on these SLSs. There is another special type of Latin square called cyclic Latin square (CLS), whose rows satisfy that $\p_{i+1}=L\p_i,\ i=1,2,\ldots,d-1$, where $\p_i$ is the $i$th row of the LS and $L$ is the left shift operator defined by $L(x_1,x_2,\ldots,x_d)=(x_2,x_3,\ldots,x_d,x_1)$. In other words, CLS is a special type of LS that is relatively easy to construct. In consideration of the computing time and the performance, we construct LS by applying a random row and column permutation on CLS in this paper. Note that COA shares the balancedness of LS, that is, each player has equal number of times appearing in each of the column. In addition, the COA ensures that each pair of players takes any given pair of columns the same number of times too.

\section{Estimates of the Shapley Value}\label{method}
\subsection{The algorithm}\label{alg}
As shown in Section \ref{pre}, a COA or an LS is a balanced fraction of all permutations---we exploit this balance to approximate the Shapley value. The estimates of the Shapley value based on LS and COA are defined as
\begin{gather*}
\widehat{\Sh}_{j}^{LS}=\frac{1}{d\nc_{LS}}\sum_{i=1}^{\nc_{LS}}\sum_{\pi\in LS_i}\Delta(\pi)_j,\quad\text{and}\\
\widehat{\Sh}_{j}^{COA}=\frac{1}{d(d-1)\nc_{COA}}\sum_{i=1}^{\nc_{COA}}\sum_{\pi\in COA_i}\Delta(\pi)_j,
\end{gather*}
for $ j=1,2,\ldots,d,$, where $LS_i$ and $COA_i$ are the $i$th randomly generated LS$(d)$ and $COA(d(d-1),d)$, and $\nc_{LS},\nc_{COA}$ are numbers of LS and COA have been used respectively.
Both designs can be used in Algorithm \ref{alg1}, where $\pi(l)$ represents the position of $l$ in the permutation $\pi$.

\begin{algorithm}[ht]
\setstretch{1.3}
\SetAlgoLined
\caption{\small{Estimating the Shapley value}}\label{alg1}
\KwIn{Given $\nc_{COA}\in \mathbb{N}$}
\KwOut{$\widehat{\Sh}_j^{COA},\ j=1,2,\ldots,d$.}
Set $c=1$ and $\widehat{\Sh}_j^{COA}=0,\ j=1,2,\ldots,d$.\\
Generating $\nc_{COA}$ COA$(d(d-1),d)$ randomly, denoted by $D_1,D_2,\ldots,D_{\nc_{COA}}$.\\
\While{$c\leq \nc_{COA}$}{
Let $\pi_{i},\ i=1,2,\ldots,d(d-1)$ be the $i$th row of $D_c$.\\
\For{$i=1,\ldots,d(d-1)$}{
 Set pre$\nu$=0.\\
   \For{$l=1,2,\ldots,d$}{
      Calculate $\Delta(\pi_i)_{\pi_i(l)}=\nu(P_{\pi_i}^{\pi_i(l)})-\text{pre}\nu$\\
      Let $\widehat{\Sh}_{\pi_i(l)}^{COA}=\widehat{\Sh}_{\pi_i(l)}^{COA}+\Delta(\pi_i)_{\pi_i(l)}$\\
      Set pre$\nu=\nu(P_{\pi_i}^{\pi_i(l)})$}
   }
   $c=c+1$\;
}
Obtain $\widehat{\Sh}_j^{COA}=\widehat{\Sh}_j^{COA}/(d(d-1)\nc_{COA}),\ j=1,2,\ldots,d$.
\end{algorithm}

 The major innovation in our algorithm is to replace the SRS by design based sampling scheme, either the COA or the LS. For the same number of permutations to be sampled, {our method yields much more accurate estimate of the Shapley values than existing methods based on SRS in most cases. More details will be deferred to Section \ref{sim} as well as Section 3 of the Supplementary Material.}
Surprisingly, we find our algorithm based on COA is even faster than SRS empirically. The possible reason is that it only needs one sampling procedure to generate a COA$(d(d-1),d)$ in the COA method, while the SRS method requires to sample $d(d-1)$ times to obtain the permutations with the same sample size of a COA.

Note that $d$ values of marginal contributions are calculated for each permutation $\pi_i$ in Algorithm \ref{alg1} while CGT algorithm requires $2d$ evaluations for each permutation. In order to compare three estimate methods fairly, the improved version of CGT algorithm is used in Section \ref{sim}.

\begin{remark}\label{remark.estnu}\rm
As discussed in Section \ref{pre.sh}, choosing different value functions can enable the algorithm to be used in different problems, such as sensitivity analysis and local model explanation. In some settings, the value of $\nu(P_{\pi_i}^{\pi_i(l)})$ in line 8 of Algorithm \ref{alg1} cannot be calculated accurately. For example, \cite{SNS2016} suggested using $\E[\var(Y|\x_{\u^c})]$ as the value function for sensitivity analysis. However, it is difficult to obtain the exact value of this value function even if the distribution of $\x$ is known. In this case, we adopt the Monte Carlo method to estimate the value function like \cite{SNS2016}, the resulting algorithm is shown in the Supplementary Material.
\end{remark}

\begin{remark}\rm
Our pseudo {code} in Algorithm \ref{alg1} {estimates the} Shapley value by
$\widehat{\Sh}_{j}^{COA}$ based on design COA$(d(d-1),d)$, which can be replaced by $\widehat{\Sh}_j^{LS}$ based on $LS(d)$. However, simulations in Section \ref{sim} show that $\widehat{\Sh}_j^{COA}$ performs better than $\widehat{\Sh}_j^{LS}$. We suggest the use of $COA$ whenever $d$ is not {prohibitively} large for it to be implemented. In the latter case, the LS is recommended. Note that the construction of COA requires $d$ to be a {prime} power, however, we can utilize the trick of adding null players to enlarge the number of players so that the new $d$ becomes a {prime} power.
\end{remark}

\subsection{Properties}\label{pro}
In this subsection, we show the properties of our estimates with COAs and LSs.
\cite{CGT2009} shown that the variance of $\widehat{\Sh}^{SRS}_j$ is given by $\var(\widehat{\Sh}^{SRS}_j)=\frac{1}{md!}\sum_{\pi\in\Pi}(\Delta(\pi)_j-\Sh_j)^2$, where $m$ is the number of sampled permutations. Variances of $\widehat{\Sh}_j^{COA}$ and $\widehat{\Sh}_j^{LS}$ are given in the following theorem.

\begin{The}\label{pro.unbias}
The estimates $\widehat{\Sh}_j^{COA}$ and $\widehat{\Sh}_j^{LS},\ j=1,2,\ldots,d$ are unbiased, and their variances are given by
\begin{eqnarray}
\var(\widehat{\Sh}_j^{COA})&=&\var(\widehat{\Sh}^{SRS}_j)+\frac{d(d-1)-1}{d(d-1)\nc_{COA}}\cov_{COA},\label{eqn:coa}\\
\var(\widehat{\Sh}_j^{LS})&=&\var(\widehat{\Sh}^{SRS}_j)+\frac{d-1}{d\nc_{LS}}\cov_{LS},\label{eqn:ls}
\end{eqnarray}
where
\[\begin{aligned}
&\cov_{COA}=\frac{1}{d![d(d-1)-1]}\sum_{i=1}^{(d-2)!}\sum_{\substack{\pi_1\neq\pi_2,\\ \pi_1,\pi_2\in COA_i}}(\Delta(\pi_1)_j-\Sh_j)(\Delta(\pi_2)_j-\Sh_j),\\
&\cov_{LS}=\frac{1}{d!T(d)}\sum_R(\Delta(\pi_1)_j-\Sh_j)(\Delta(\pi_2)_j-\Sh_j), \end{aligned}\]
$COA_i,i=1,2,\ldots,(d-2)!$ are partitions of $\Pi$ by \cite{YSX2021}, $T(d)=d!(1/2-1/3!+\cdots+(-1)^d/d!)$ and $R$ is the restricted space of $d!T(d)$ pairs $(\pi_1,\pi_2)$ that {have} no common value at any position.
\end{The}

\begin{remark}\rm
Theorem \ref{pro.unbias} shows that $\widehat{\Sh}_j^{COA}$ and $\widehat{\Sh}_j^{LS}$ are unbiased, and variances of these two estimates converge to 0 as $n\to \infty$. Besides, the consistent property can be easily obtained with their variances and Chebyshev's inequality, i.e.
$\lim_{n\to\infty}P(|\widehat{\Sh}_j-\Sh_j|>\epsilon)=0,\ \forall\epsilon>0,\ j=1,2,\ldots,d.$
\end{remark}

$\widehat{\Sh}_j^{COA}$ has smaller variance than that of $\widehat{\Sh}^{SRS}_j$ if and only if $\cov_{COA}<0$. Similarly, $\var(\widehat{\Sh}_j^{LS})<\var(\widehat{\Sh}^{SRS}_j)$ if and only if $\cov_{LS}<0$. Intuitively, such negativity conditions are very natural to hold since the designs LS and COA tries to fill the space of the permutations and hence the assembled permutations are designed to be as distinct as possible. However, it can not be proven that $\widehat{\Sh}_j^{COA}$ and $\widehat{\Sh}_j^{LS}$ have smaller variances than $\widehat{\Sh}^{SRS}_j$ in general case. {The simulation under a manufacturing system model shows that $\var(\widehat{\Sh}_j^{LS})$ or $\var(\widehat{\Sh}_j^{COA})$ may slightly larger than $\var(\widehat{\Sh}^{SRS}_j)$ for a few specific variable, although the sum of variances of all estimates is always the smallest for $\widehat{\Sh}^{COA}$ and the largest for $\widehat{\Sh}^{SRS}$. More details will be deferred to Section 3 of the Supplementary Material.} While, in many cases, we find that the estimates based on COA or LS perform better
than SRS both empirically and theoretically. We provide two examples below for demonstrations. The first example is a symmetric voting game and the second example is a linear model with normal distribution inputs.

\begin{Exa}\label{exa.svoting}
{\rm Let $N=\{1,2,\ldots,d\}$ be the $d$ players with even $d$. For any $\u\subset N$, the value function of the symmetric voting game be
\[\begin{aligned}
\nu(\u)=\left\{\begin{array}{ll}
1, &\text{if}\ |\u|>d/2;\\
0, &\text{otherwise}.\\
\end{array}\right.
\end{aligned}\]
It is straightforward to show that $\Sh_j=1/d,\ j=1,2,\ldots,d$, and
$$
\Delta(\pi)_j=\left\{\begin{array}{ll}
1, &\text{if}\ |P_\pi^j|=1+d/2;\\
0, &\text{otherwise}.\\
\end{array}\right.
$$
First, we have $\var(\widehat{\Sh}^{SRS}_j)=(d-1)/(md^2)$. Since COAs are balanced, each player appears at each position $(d-1)$ times in a COA$(d(d-1),d)$, which implies that for $i=1,2,\ldots,(d-2)!$,
\[\begin{aligned}
&\sum_{\substack{\pi_1\neq\pi_2,\\ \pi_1,\pi_2\in COA_i}}(\Delta(\pi_1)_j-\Sh_j)(\Delta(\pi_2)_j-\Sh_j)\\
=&(d-1)(d-2)\left(1-\frac{1}{d}\right)^2+2(d-1)^3\left(1-\frac{1}{d}\right)\left(-\frac{1}{d}\right)+(d-1)^2[(d-1)^2-1]\left(-\frac{1}{d}\right)^2\\
=&-\frac{(d-1)^2}{d}<0,
\end{aligned}\]
plugging this result into (\ref{eqn:coa}), we have $\var(\widehat{\Sh}_j^{COA})=0$.

For the LS based estimate, with the definition of $R$, we have
\[\begin{aligned}
\cov_{LS}=&\frac{1}{d!T(d)}\Bigg[(d-1)!T(d)\left(1-\frac{1}{d}\right)\left(-\frac{1}{d}\right)+\left(d!-(d-1)!\right)\frac{T(d)}{d-1}\left(1-\frac{1}{d}\right)\left(-\frac{1}{d}\right)\\
&+\left(d!-(d-1)!\right)\left(1-\frac{1}{d-1}\right)T(d)\left(-\frac{1}{d}\right)^2\Bigg]\\
=&-1/d^2<0,
\end{aligned}\]
plugging this result into (\ref{eqn:ls}), we also have $\var(\widehat{\Sh}_j^{LS})=0$. In other words, we have $\var(\widehat{\Sh}_j^{COA})=\var(\widehat{\Sh}_j^{LS})=0<\var(\widehat{\Sh}^{SRS}_j)=(d-1)/(md^2)$ for the symmetric voting game.
}
\end{Exa}

\begin{Exa}\label{exa.linear}
\rm{ Let $h(\textbf{x})=\beta_0+\beta^T\textbf{x}$ with $\textbf{x}\sim N(\mu,\Sigma),\ \Sigma=(\Sigma_{ij})_{d\times d}$, where
$$
\Sigma_{ij}=\left\{\begin{array}{ll}
{\sigma_i^2,} &{\text{if}\ i=j};\\
{\sigma_{i,j},} &{\text{if}\ j\leq 2t,\ \text{and}\ j=i+1,\ j=0\ (\text{mod}\ 2)\ \text{or}\ j=i-1,\ i=0\ (\text{mod}\ 2)};\\
{0,} &{\text{others}}.
\end{array}\right.
$$
and $0\leq t\leq\lfloor d/2\rfloor$. $\lfloor x\rfloor$ denotes the largest integer not exceeding $x$. \cite{OP2017} showed that when $h(\textbf{x})=\beta_0+\beta^T\textbf{x}$ with $\textbf{x}\sim N(\mu,\Sigma)$, where $\Sigma$ has full rank, the expression of $\Sh_j$ is
$$\Sh_j=\frac{1}{d}\sum_{\u\subseteq-j}{(d-1)\choose{|\u|}}^{-1}\frac{\cov(x_j,\textbf{x}_{\u^c}^T\beta_{\u^c}|\textbf{x}_{\u})^2}{\var(x_j|\textbf{x}_{\u})}.$$
From their proof, it can be seen that $\Delta(\pi)_j=\cov(x_j,\textbf{x}_{-P_{\pi}^j}^T\beta_{-P_{\pi}^j}|\textbf{x}_{P_{\pi}^j})^2\Big/\var(x_j|\textbf{x}_{P_{\pi}^j})$. When $j=0\ (\text{mod}\ 2)$ and $j\leq 2t$, it can be calculated that there are two kinds of $\Delta(\pi)_j$.\\
Case 1: $j-1\in P_{\pi}^j$. In this case $\Delta(\pi)_j=\beta_j^2(\sigma_j^2-\sigma_j^{-2}\sigma_{j-1,j})\triangleq g_1$.\\
Case 2: $j-1\notin P_{\pi}^j$. In this case $\Delta(\pi)_j=\sigma_j^{-2}(\sigma_{j-1,j}\beta_{j-1}+\sigma_j^2\beta_j)^2\triangleq g_2$.\\
For any $j=1,2,\ldots,d$, there are $d(d-1)/2$ permutations in a COA such that $j\in P_{\pi}^j$. Thus, for $i=1,2,\ldots,(d-2)!$,
\[\begin{aligned}
&\sum_{\substack{\pi_1\neq\pi_2,\\ \pi_1,\pi_2\in COA_i}}(\Delta(\pi_1)_j-\Sh_j)(\Delta(\pi_2)_j-\Sh_j)\\
=&\frac{d(d-1)}{2}\left(\frac{d(d-1)}{2}-1\right)\left(\frac{1}{2}g_1-\frac{1}{2}g_2\right)^2+\frac{d(d-1)}{2}\left(\frac{d(d-1)}{2}-1\right)\left(-\frac{1}{2}g_1+\frac{1}{2}g_2\right)^2\\
&+\frac{d^2(d-1)^2}{2}\left(\frac{1}{2}g_1-\frac{1}{2}g_2\right)\left(-\frac{1}{2}g_1+\frac{1}{2}g_2\right)\\
=&-d(d-1)\left(\frac{1}{2}g_1-\frac{1}{2}g_2\right)^2<0,
\end{aligned}\]
this together with $\var(\widehat{\Sh}^{SRS}_j)=(g_1-g_2)^2/(4t)$ leads to the result $\var(\widehat{\Sh}_j^{COA})=0$ based on the calculation of (\ref{eqn:coa}). On the other hand, we can verify that $\cov_{LS}=0$. So in summary, we have $\var(\widehat{\Sh}_j^{COA})=0<\var(\widehat{\Sh}^{SRS}_j)=\var(\widehat{\Sh}_j^{LS})$} here.
\end{Exa}

In Section \ref{pre.sh}, we {stated} four axioms which suggest the Shapley value is a reasonable measure in cooperative games. \cite{SNS2016} showed the Shapley value always decompose the total variance and allocate to each input regardless of the dependence among inputs or the structure of the output function, whereas other commonly used variance-based sensitivity measure do not. In fact, the efficient property of axiom 1 always holds for $\widehat{\Sh}_j^{COA}$ and $\widehat{\Sh}_j^{LS}$.

\begin{Pro}\label{pro.eff}
$\widehat{\Sh}_j^{COA}$ and $\widehat{\Sh}_j^{LS}$ are efficient, i.e. $\sum_{j=1}^d\widehat{\Sh}_j^{COA}=\sum_{j=1}^d\widehat{\Sh}_j^{LS}=\nu(\{1:d\}).$
\end{Pro}

We now show that these two estimates equal the true Shapley value in some special cases, which is mainly guaranteed by the balance property of $LS$ and $COA$, i.e., each player has equal number of times appearing in each of the column, and doesn't hold for SRS.

\begin{The}\label{pro.real}
In any of the following cases, $\widehat{\Sh}_j^{COA}=\Sh_j$. And in the first two cases, $\widehat{\Sh}_j^{LS}=\Sh_j$.\\
(1) $\forall \u\subset\{1:d\}\backslash\{j\},\ \nu(\u\cup\{j\})=\nu(\u)+\nu(\{j\}). $\\
(2) $\Delta(\pi)_j$ is a function of the position of $\{j\}$ in permutation $\pi$ for all $\pi\in\Pi$.\\
(3) $\Delta(\pi)_j$ is a function of $I_{i,j},\ i\in\{1:d\}\backslash\{j\}$, where $I_{i,j}$ is the pseudo factor with two levels, 1 and $-1$, indicating whether or not the position of $i$ is before it of $j$ in a permutation.
\end{The}

It can be seen that $\Delta(\pi)_j,\ j=1,2,\ldots,d$ in Example \ref{exa.svoting} satisfy case (2) and $\Delta(\pi)_j,\ j=1,2,\ldots,d$ in Example \ref{exa.linear} satisfy case (3) of Theorem \ref{pro.real}.

\begin{figure}[!ht]
	\centering
	\begin{tabular}{cc}
		\includegraphics[totalheight=2in, width=3in]{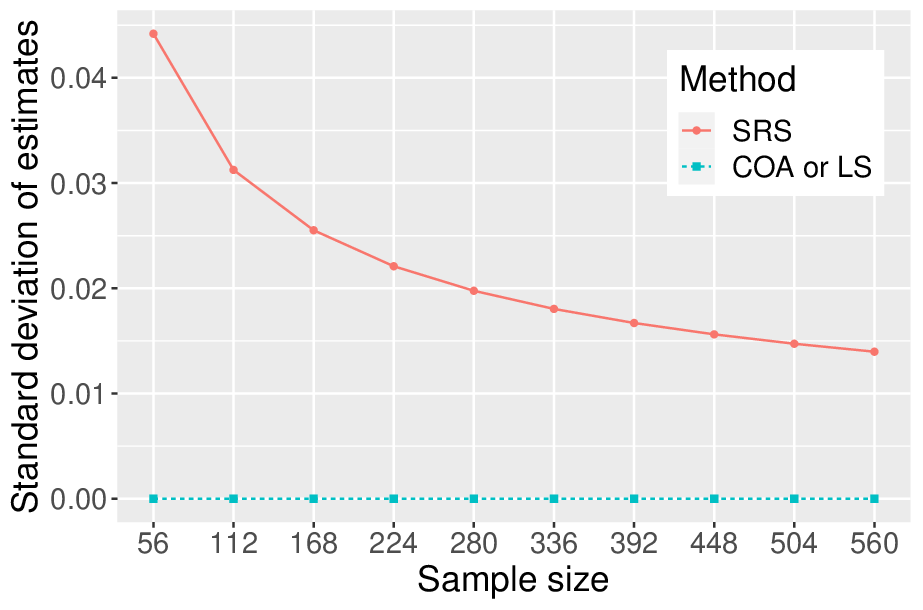} &
		\includegraphics[totalheight=2in, width=3in]{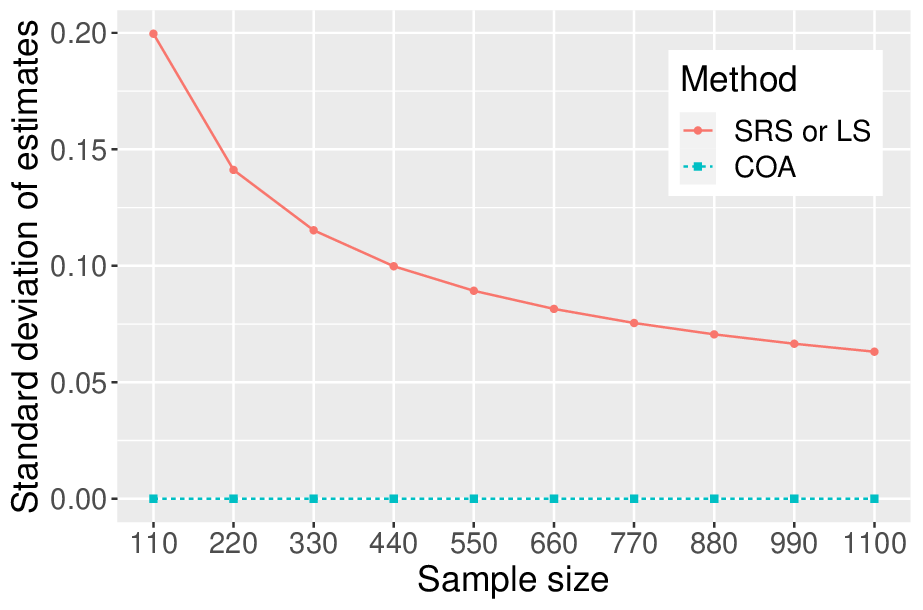}\\
		(a) Example \ref{example3}: a voting game with $d=8$&
		(b) Example \ref{example4}: sensitivity analysis with $d=11$
	\end{tabular}
	\caption{Comparisons of the standard deviations of different Shapley value estimates for Examples \ref{example3} and \ref{example4}.
	}\label{fig.var}
\end{figure}

\begin{Exa}\label{example3}
\rm{(Example \ref{exa.svoting} continued) For a symmetric voting game, $\Delta(\pi)_j$ is a function of the position of $\{j\}$ in permutation $\pi$ for all $\pi\in \Pi$, which satisfies case (2) of Theorem \ref{pro.real}. Thus, there is always $\widehat{\Sh}_j^{COA}=\widehat{\Sh}_j^{LS}=\Sh_j=0.125$. For SRS method, recall we have $\var(\widehat{\Sh}^{SRS}_j)=(d-1)/(md^2)$ for $j=1,2,\ldots,d$. Figure \ref{fig.var}(a) shows the standard deviations of these estimates with $d=8$.
}
\end{Exa}

\begin{Exa}\label{example4}
\rm{(Example \ref{exa.linear} continued)
Consider the model and distribution of inputs in Example \ref{exa.linear}. When $j=0\ (\text{mod}\ 2)$ and $j\leq 2t$, $\Delta(\pi)_j=g_1\textbf{1}_{\{I_{j-1,j}=1\}}+g_2\textbf{1}_{\{I_{j-1,j}=-1\}}$, which satisfies the case (3) in Theorem \ref{pro.real}, where $g_1$ and $g_2$ are defined in Example \ref{exa.linear}. Similar conclusions of other cases of $j$ can also be shown. Thus, $\widehat{\Sh}_j^{COA}=\Sh_j=(g_1+g_2)/2$ for $j=1,2,\ldots,d$. Recall $\var(\widehat{\Sh}^{SRS}_j)=\var(\widehat{\Sh}_j^{LS})=(g_1-g_2)^2/(4m)$. See Figure \ref{fig.var}(b) for the comparison of the standard deviations with $d=11,\beta_1=1,\beta_2=2,\sigma_{1,2}=-0.5$ and $\sigma_2^2=0.8$.
}
\end{Exa}

Sometimes various value functions are used to measure the contribution of the players at a same time. Thus, a linear combination of multiple Shapley values needs to be calculated.
\begin{Pro}\label{var.linear}
Suppose that $\nu_1$ and $\nu_2$ are two value functions such that $\var(\widehat{\Sh}_j(\nu_1))=\var(\widehat{\Sh}_j(\nu_2))=0$, where $\widehat{\Sh}$ is obtained with SRS, COA or LS method. Let  $\nu$ be any linear combination of $\nu_1$ and $\nu_2$. Then, we have $\var(\widehat{\Sh}_j(\nu))=0$.
\end{Pro}

With Theorem \ref{pro.real} and Proposition \ref{var.linear}, a linear space of value functions such that $\widehat{\Sh}_j^{COA}=\Sh_j$ can be obtained.
\begin{Cor}
If $\nu$ belongs to the linear space $\mathcal{V}=\{\nu\mid\text{for\ any\ permutation}\ \pi: \nu(P_{\pi}^j\cup\{j\})-\nu(P_{\pi}^j)=\sum_{i\in\{j\}^c}h_i(I_{ij}(\pi))\}$, where $h_i$ is a real function and $I_{i,j}(\pi)$ is the pseudo factor with two levels, 1 and $-1$, indicating whether or not the position of $i$ is before it of $j$ in $\pi$, then $\var(\widehat{\Sh}^{COA}_j(\nu))=0$.
\end{Cor}

\subsection{Toward Arbitrary Numbers of Players}\label{flexible}
We have established that $\widehat{\Sh}^{COA}$ is the most accurate estimate of the Shapley value among the three designs. We noted earlier that it lacks some flexibility in requiring that $d$ is a prime power. We shall modify the COA design to accommodate arbitrary values of $d$. As mentioned in Section \ref{pre.oofa}, we can add null players so that the total number of players, say $d^*$, become a prime power. Those null players do not contribute the value function and hence, based on axiom 3, do not affect the calculation of the Shapley values of existing players. For a new sequence $\pi^*$ with $d^*$ players, the marginal contribution of player $j$ in this sequence is simply $\Delta\left(\pi^*\backslash \overline{\d^c}\right)_j$, where $\overline{\d^c}=\{(d+1):d^*\}$. Hence we shall define the modified COA estimate as

\begin{equation}\label{eq.gcoa}
\widehat{\Sh}_j^{COA^*}=\frac{1}{d^*(d^*-1)\nc_{COA^*}}\sum_{i=1}^{\nc_{COA^*}}\sum_{\pi^*\in COA_i^*}\Delta\left(\pi^*\backslash \overline{\d^c}\right)_j,\quad j=1,2,\ldots,d,
\end{equation}
where $COA_i^*$ is the $i$th randomly generated $COA(d^*(d^*-1),d^*)$, $i=1,2,\ldots,\nc_{COA^*}$. The following corollary provides the property of $\widehat{\Sh}_j^{COA^*}$. Section \ref{sec:911} provides a real data demonstration for this new estimate.
\begin{Cor}\label{coro.gcoa}
The estimates $\widehat{\Sh}_j^{COA^*}$, $j=1,2,\ldots,d$ are unbiased, and their variances are given by
$$
\var(\widehat{\Sh}_j^{COA^*})=\var({\widehat{\Sh}^{SRS}_j})+\frac{d^*(d^*-1)-1}{d^*(d^*-1)\nc_{COA^*}}\cov_{COA^*},
$$
where
$$\cov_{COA^*}=\frac{1}{d^*![d^*(d^*-1)-1]}\sum_{i=1}^{(d^*-2)!}\sum_{\substack{\pi^*_1\neq\pi^*_2,\\ \pi^*_1,\pi^*_2\in COA^*_i}}(\Delta(\pi^*_1\backslash \overline{\d^c})_j-\Sh_j)(\Delta(\pi^*_2\backslash \overline{\d^c})_j-\Sh_j).
$$
\end{Cor}

\section{Simulations}\label{sim}

In this section, we compare $\widehat{\Sh}^{SRS}_j$, $\widehat{\Sh}^{StrRS}$, $\widehat{\Sh}_j^{COA}$ and $\widehat{\Sh}_j^{LS}$ as defined earlier in this paper under an airport game, where $\widehat{\Sh}^{SRS}_j$ is obtained by improved CGT algorithm mentioned in Section \ref{alg}, $\widehat{\Sh}^{StrRS}$ is obtained by the algorithm in \cite{vHHL2018}, $\widehat{\Sh}_j^{COA}$ is derived by Algorithm \ref{alg1}, and $\widehat{\Sh}_j^{LS}$ is a counterpart of $\widehat{\Sh}_j^{COA}$ by replacing the COA design by LS design in Algorithm \ref{alg1}.  We adopt equation (\ref{coa}) to generate COA and construct LS  based on a random row and column permutation on CLS. Additional simulations for the voting game, sensitivity analysis and local model explanation are deferred to the Supplementary Material due to limit of space.

\begin{figure}[!ht]
\centering
\begin{tabular}{cc}
\includegraphics[totalheight=2in, width=3.1in]{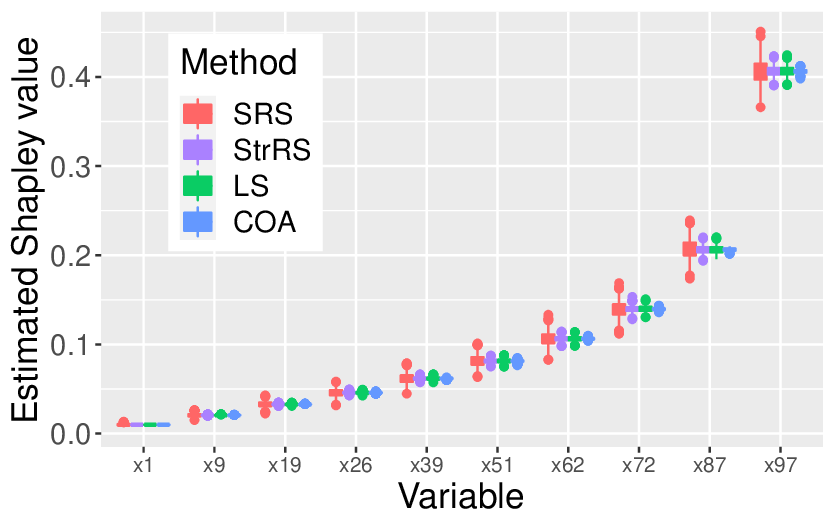} &
\includegraphics[totalheight=2in, width=3.1in]{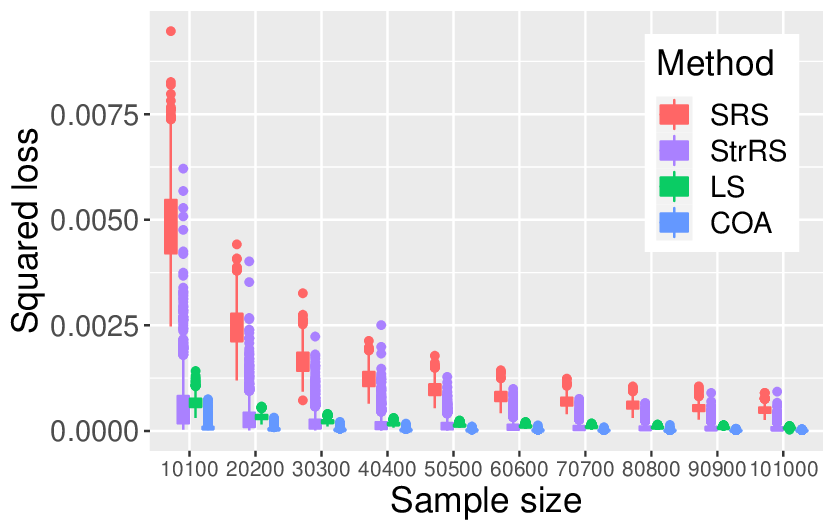}\\
 (a) Estimates of $\Sh$ with $m=10100$ &(b) Squared loss of $\widehat{\Sh}$ with 10 sample sizes
\end{tabular}
\caption{Estimates and squared loss with four methods under an airport game with $d=101$. The boxplots are based on 1000 realizations.
}\label{figure.air}
\end{figure}

The airport game is a cost sharing problem proposed by \cite{LO1973}. The goal in its original form is to distribute the cost of a landing strip among users who need runways of different lengths.
Let $N=\{1:101\}$ be the user set, and the value function of $\u\subset N$ be $\nu(\u)=\max_{i\in \u}w_i$, where $\w=(w_1,w_2,\ldots,w_{101})=(\one_8^T,\two_{10}^T,\three_7^T,\four_{13}^T,\five_{12}^T,\six_{11}^T,\seven_{10}^T,\eight_{15}^T,\nine_{10}^T,\ten_{5}^T)$
is the vector of runway length required by 101 users, and $\k_l$ is an $l\times1$ vector with all elements $k$. \cite{CGT2008} proved that the Shapley value for such airport game is
$$
Sh_j=\sum_{k=1}^{j}\frac{w_k-w_{k-1}}{d+1-k},\quad \text{for}\ j=1,\ldots,d,
$$
where $w_0=0$.
Boxplots of estimates and squared loss with four methods have been shown in Figure \ref{figure.air}.  The boxplots are based on 1000 realizations.

It can be seen that the performance of $\widehat{\Sh}^{COA}$, $\widehat{\Sh}^{LS}$ and $\widehat{\Sh}^{StrRS}$ are much better than that of $\widehat{\Sh}^{SRS}$ in both estimation accuracy and variance. Moreover, $\widehat{\Sh}^{COA}$ is more accurate than $\widehat{\Sh}^{LS}$, and both of them are more robust than $\widehat{\Sh}^{StrRS}$.

We also compare the computation time of $\widehat{\Sh}^{SRS}$, $\widehat{\Sh}^{StrRS}$, $\widehat{\Sh}^{LS}$ and $\widehat{\Sh}^{COA}$, and the result are shown in Table \ref{tab.time}, including ratios of computation time and the average CPU seconds of $\widehat{\Sh}^{COA}$. The machine we used is  Lenovo SR860, 240 cores. It can be seen that $\widehat{\Sh}^{COA}$ is always the fastest, and $\widehat{\Sh}^{StrRS}$ is much more time consuming than other estimates.
\begin{table}[htbp]
\centering
\caption{Ratios of average CPU seconds among 1000 realizations of four methods to the COA method under an airport game with $d=101$ and different sample sizes. The CPU seconds for the COA method is given in the parentheses.}
\label{tab.time}\renewcommand\tabcolsep{2.1pt}
\begin{tabular}{c|cccccccccc}
\hline
\multirow{2}{*}{Method} & \multicolumn{10}{c}{Sample size}                                                                                                                                                                                                                                                                                                                                                                                                                                                                                                                             \\ \cline{2-11}
                        & 10100                                               & 20200                                               & 30300                                               & 40400                                               & 50500                                               & 60600                                               & 70700                                               & 80800                                                & 90900                                                & 101000                                               \\ \hline
SRS                     & 1.39                                                & 1.27                                                & 1.25                                                & 1.21                                                & 1.19                                                & 1.20                                                 & 1.19                                                & 1.15                                                 & 1.16                                                 & 1.14                                                 \\ \hline
StrRS                   & 5.31                                                & 5.36                                                & 5.41                                                & 5.49                                                & 5.54                                                & 5.41                                                & 5.44                                                & 5.27                                                 & 5.32                                                 & 5.23                                                 \\ \hline
LS                      & 1.37                                                & 1.30                                                 & 1.31                                                & 1.31                                                & 1.31                                                & 1.32                                                & 1.32                                                & 1.30                                                  & 1.30                                                  & 1.29                                                 \\ \hline
COA                     & \begin{tabular}[c]{@{}c@{}}1.00\\ (14.88)\end{tabular} & \begin{tabular}[c]{@{}c@{}}1.00\\ (29.25)\end{tabular} & \begin{tabular}[c]{@{}c@{}}1.00\\ (42.62)\end{tabular} & \begin{tabular}[c]{@{}c@{}}1.00\\ (52.22)\end{tabular} & \begin{tabular}[c]{@{}c@{}}1.00\\ (69.59)\end{tabular} & \begin{tabular}[c]{@{}c@{}}1.00\\ (82.77)\end{tabular} & \begin{tabular}[c]{@{}c@{}}1.00\\ (96.84)\end{tabular} & \begin{tabular}[c]{@{}c@{}}1.00\\ (111.33)\end{tabular} & \begin{tabular}[c]{@{}c@{}}1.00\\ (125.32)\end{tabular} & \begin{tabular}[c]{@{}c@{}}1.00\\ (139.54)\end{tabular}\\ \hline
\end{tabular}
\end{table}

\section{Applications in Real Data}\label{realdata}
In this section, we analysis the real data sets using the proposed estimates, including the 9/11 terrorist network and the connectome of C. elegans nervous system. Social network analysis are carried out to these two data sets to identify the important members, where the value function is defined as $\nu(\u)=1$ if the subnetwork composed of $\u$ is connected and $\nu(\u)=0$ otherwise for any coalition $\u$.

\subsection*{(1) 9/11 terrorist network}\label{sec:911}
The 9/11 terrorist attack on September 11, 2001 {was} a serious terrorist attack by the Wahhabi Islamist terrorist group Al-Qaeda against the United States.
A major problem in counterterrorism practice is to identify the key members in a terrorist attack by analyzing the social network, for which the Shapley value provides a viable solution. \cite{K2002} mapped a network around the 19 dead pilots and hijackers in the 9/11 terrorist attack using the public data. This terrorist network were expanded by \cite{vHHL2018} to a network of 69 members including pilots, hijackers and accomplices. The importance of these 69 members could be ranked based on their importance as measure by a Shapley value.

The real Shapley value cannot be calculated since {$2^{69}$} is too large. \cite{vHHL2018} has verified that their results of {the set of top 15 members are} the same when the SRS sample size is increased from 20000 to 500000. Here we further verify their list with a SRS sample of size 22.356 millions, and we regard this $\widehat{\Sh}^{SRS}$ as the pseudo Shapley value to calculate the squared loss and identify the key members. Hereafter we show the pseudo top 20 members and their pseudo Shapley value in Table \ref{911.pseudo}.

\begin{table}[!ht]
\centering
\caption{The top 20 key members of 9/11 terrorist network and their pseudo Shapley value.}
\label{911.pseudo}\renewcommand\tabcolsep{2.5pt}
\renewcommand\arraystretch{0.85}
\begin{tabular}{ccc|ccc}
\hline
Rank & Member                 & Pseudo $\Sh$     & Rank & Member                    & Pseudo $\Sh$     \\ \hline
1    & Mohamed Atta           & 0.114013 & 11   & Hamza Alghamdi            & 0.008895 \\
2    & Essid Sami Ben Khemais & 0.111274 & 12   & Fayez Ahmed               & 0.008783 \\
3    & Hani Hanjour           & 0.110679 & 13   & Marwan Al-Shehhi          & 0.004566 \\
4    & Djamal Beghal          & 0.107299 & 14   & Satam Suqami              & 0.003696 \\
5    & Khalid Almihdhar       & 0.107244 & 15   & Saeed Alghamdi            & 0.003684 \\
6    & Mahmoun Darkazanli     & 0.106564 & 16   & Imad Eddin Barakat Yarkas & 0.001604 \\
7    & Zacarias Moussaoui     & 0.101142 & 17   & Abdul Aziz Alomari        & 0.001560 \\
8    & Nawaf Alhazmi          & 0.099555 & 18   & Ziad Jarrah               & 0.001526 \\
9    & Ramzi Bin al-Shibh     & 0.098460 & 19   & Ahmed Alghamdi            & 0.001466 \\
10   & Raed Hijazi            & 0.094862 & 20   & Said Bahaji               & 0.001451\\ \hline
\end{tabular}
\end{table}

\begin{figure}[!ht]
\centering
\begin{tabular}{cc}
\includegraphics[totalheight=2.1in, width=3.1in]{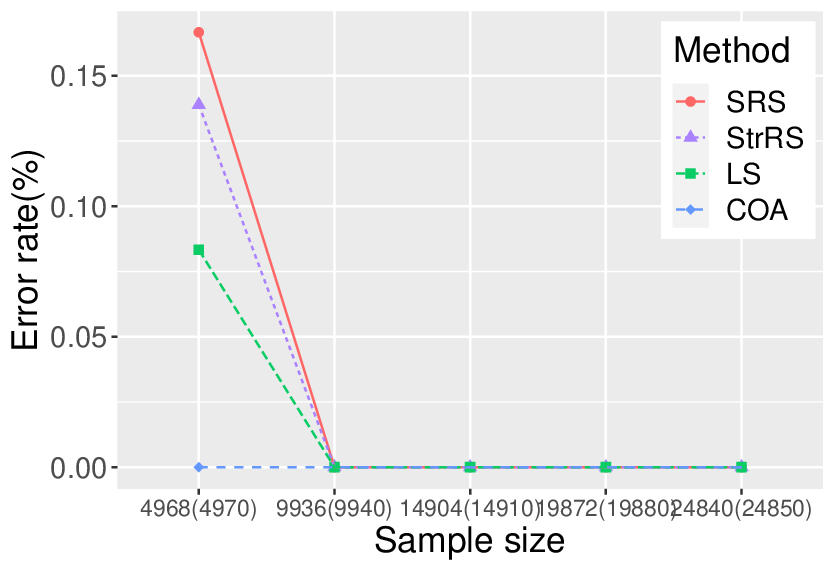} &
\includegraphics[totalheight=2.1in, width=3.1in]{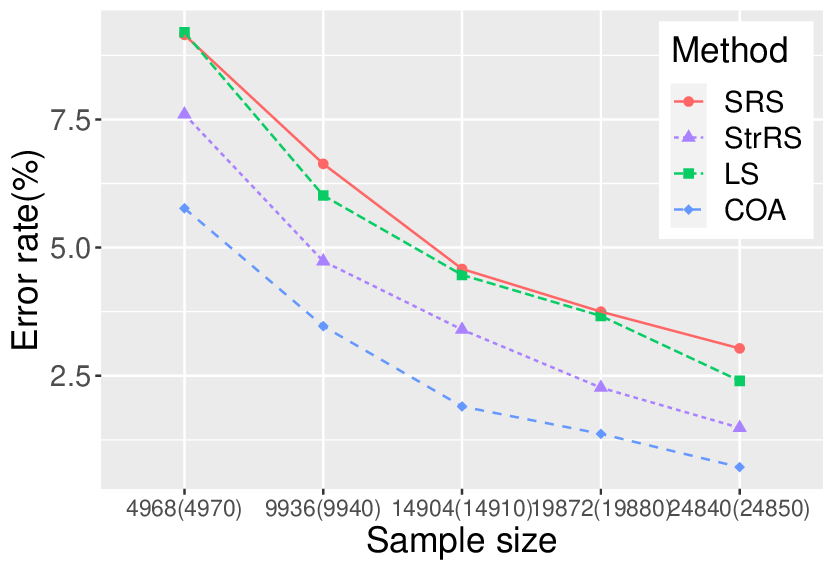}\\
(a) Error rates of top 12 key members&
(b) Error rates of top 20 key members
\end{tabular}
\caption{Error rates of the set of key members with four methods under 9/11 network with $d=69$. The error rates are simulated with 300 realizations.
}\label{fig.errorrate}
\end{figure}

For the following simulations, we replicate 300 times with 5 different sample sizes. Note that the estimate based on the COA is obtained by Equation (\ref{eq.gcoa}) with $d^*=71$ since 69 is not a prime power. The sample sizes for $\widehat{\Sh}^{COA}$ are $71\times70,71\times70\times2,\ldots,71\times70\times5$ and for other three estimates are $\lfloor71\times70/69\rfloor\times69,\lfloor71\times70/69\rfloor\times69\times2,\ldots,\lfloor71\times70/69\rfloor\times69\times5$. We first consider the accuracy of estimating the set of key members.
\cite{vHHL2018} showed that the set of top 10 key members can be correctly estimated using $\widehat{\Sh}^{StrRS}$ with 20000 samples. In the simulation, we find that no matter which method is used with {any of the sample sizes we used from 4968 to 24840}, the set of top 10 key members can always be obtained correctly. This is caused by the great difference between the Shapley values of the 10th member and the 11th member. We note that the differences between the Shapley values of the 12th and the 13th key members, and between the Shapley values of the 20th and the 21st key members are significant and not particularly large, which prompts us to compare the performance of the four methods in estimating the set of the top 12 and the top 20 key members.
Figure \ref{fig.errorrate} shows the average error rates, the proportion of the number of members that are incorrectly estimated into the set of top 12 or top 20 key members, among 300 replications of the four methods. It can be seen that, the error rates of $\widehat{\Sh}^{COA},\widehat{\Sh}^{LS}$ and $\widehat{\Sh}^{StrRS}$ converge to 0 faster than that of $\widehat{\Sh}^{SRS}$. Moreover, it takes more than twice time for calculating $\widehat{\Sh}^{StrRS}$ comparing with that of $\widehat{\Sh}^{LS}$ or $\widehat{\Sh}^{COA}$.
Thus, although $\widehat{\Sh}^{StrRS}$ seems to perform better than $\widehat{\Sh}^{LS}$ for estimating the set of top 20 key members, the accuracy of $\widehat{\Sh}^{LS}$ may be higher than that of $\widehat{\Sh}^{StrRS}$ when the computation time is the same. It can be seen that $\widehat{\Sh}^{COA}$ performs the best for estimating the top 20 key members.

\begin{figure}[!ht]
\centering
\begin{tabular}{cc}
\includegraphics[totalheight=2.2in, width=3.1in]{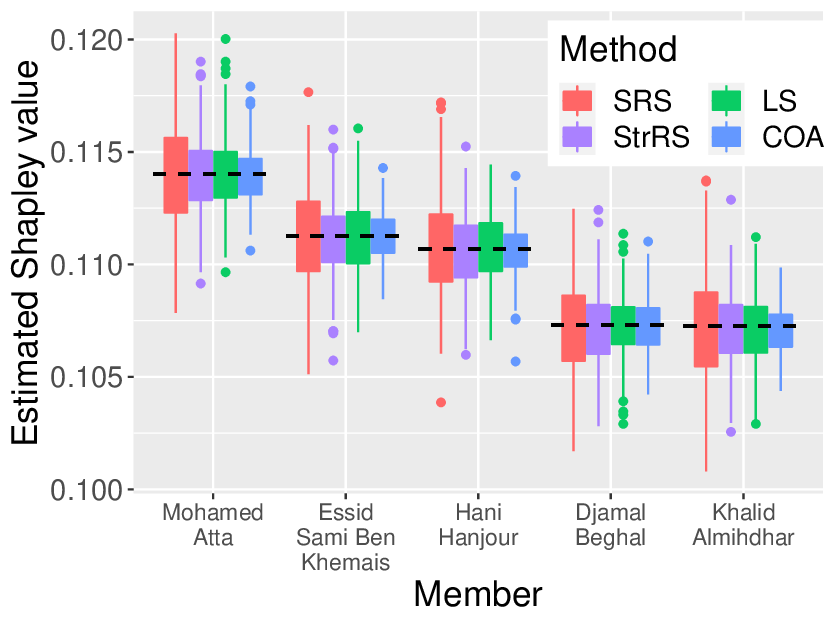} &
\includegraphics[totalheight=2.2in, width=3.1in]{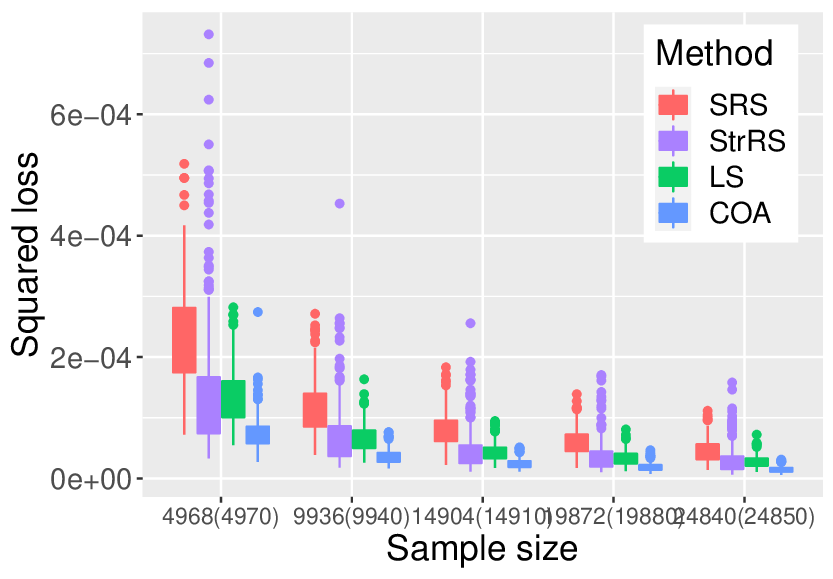}\\
\begin{tabular}{l}(a) Estimates of top 5 key members\\~~~~ with sample size $m=19872(19880)$\end{tabular}&
(b) Squared loss of $\widehat{\Sh}$ with 5 sample sizes
\end{tabular}
\caption{Estimates and squared loss with four methods under 9/11 network $d=69$. The boxplots are based on 300 realizations.
}\label{fig.mse.social}
\end{figure}

To compare the variance of estimates, the boxplots of the estimated Shapley values with multiple realizations of each estimate with sample size $m = 19872$ and $m^*=19880$ are given. In Figure \ref{fig.mse.social} (a), the dashed line is the pseudo Shapley value. It can be seen that there are significant differences in the variances of these estimates. The variances of $\widehat{\Sh}^{LS}$ and $\widehat{\Sh}^{StrRS}$ are less than that of $\widehat{\Sh}^{SRS}$, and the variance of $\widehat{\Sh}^{COA}$ is the smallest. Figure \ref{fig.mse.social} (b) gives the boxplots of squared loss with four methods and different sample size, and it is shown that $\widehat{\Sh}^{COA}$ is the best.

\subsection*{(2) Connectome of C. elegans nervous system}
The nematode C. elegans is a small invertebrate whose nervous system, general anatomy, and normal development are all (comparatively) extremely simple and reproducible, and have all been well characterised. The connectome of C. elegans nervous system is defined by thousands of synapses and gap junction allowing one to investigate neural circuits involved in regulation of innate immunity or behavior. The data set we used can be downloaded from \url{https://github.com/3BIM20162017/CElegansTP}. {There are 299 neurons and 2280 connectome pathway, which form two disjoint connected network with 279 neurons and 20 neurous respectively. To identify the important neurons in the larger connected network, we estimate the Shapley value with the value function $\nu(u) = 1$ if the subnetwork composed of $u$ is connected and $\nu(u) = 0$ otherwise.
Note that, the number of neurons to be measured, $d=279$, is too large to use the COA method, since calculating  $\widehat{\Sh}^{COA}$ with the smallest sample size requires about two days.} Besides, the simulation in Section \ref{sim} shows that the StrRS method is much more time-consuming than other methods. Thus, we only compare the performance of $\widehat{\Sh}^{LS}$ and $\widehat{\Sh}^{SRS}$.

\begin{figure}[!ht]
\centering
\begin{tabular}{cc}
\includegraphics[totalheight=2.2in, width=3.1in]{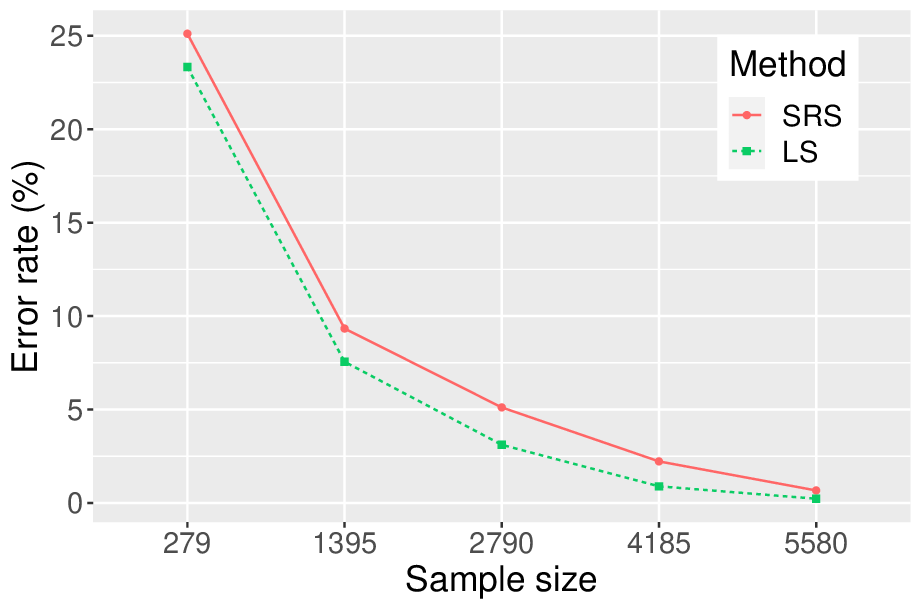} &
\includegraphics[totalheight=2.2in, width=3.1in]{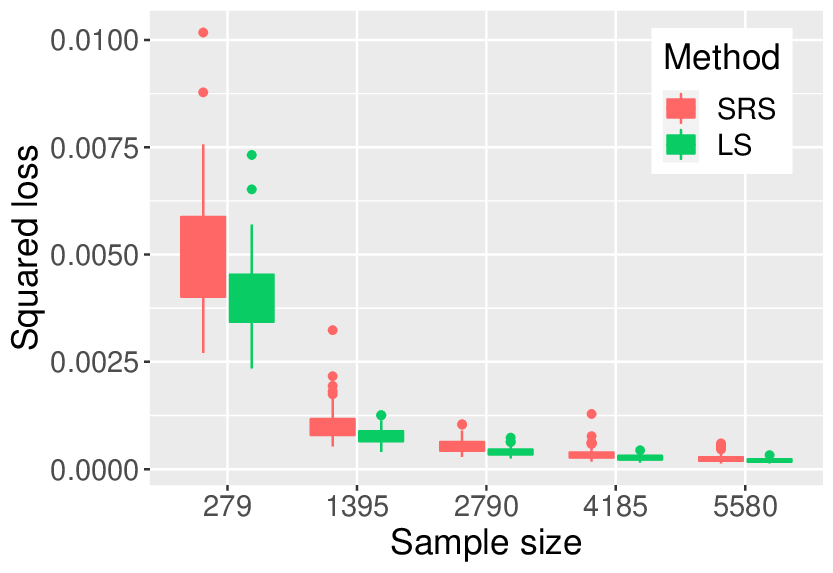}\\
(a) Error rates of top 5 key neurons&
(b) Squared loss of $\widehat{\Sh}$ with 5 sample sizes
\end{tabular}
\caption{Error rates and squared loss with LS and SRS methods under C. elegans neuvous system with $d=279$. The boxplots are based on 90 realizations.
}\label{fig.mse.nn}
\end{figure}

Figure \ref{fig.mse.nn} shows the boxplots of the error rates of top 5 key neurons and the squared loss of $\widehat{\Sh}$ with 5 sample sizes among 90 replications of the SRS and the LS methods. It can be seen that  $\widehat{\Sh}^{LS}$ performs better than $\widehat{\Sh}^{SRS}$, which suggests that estimating the Shpaley value based on the LS is a good choice when $d$ is large.

 The above simulations and applications show that $\widehat{\Sh}^{StrRS}$ is much more time-consuming than other methods, and estimates based on OofA experimental designs perform better than SRS in both squared loss and variance of estimates. Moreover, $\widehat{\Sh}^{COA}$ performs better than $\widehat{\Sh}^{LS}$ whenever a COA exists, that is when $d$ is a prime power.

\section{Concluding Remarks}\label{conc}
Among the existing methods for estimating the Shapley value, algorithms proposed by \cite{CGT2009} and \cite{SNS2016} are the most general method, which obtain the estimate with simple random sampling (SRS) method of permutations. In this paper, we consider using two types of OofA experimental designs as alternative sampling schemes.
They are Latin square (LS) and component orthogonal array (COA). These estimates are unbiased, and even equal the true Shapley values in some special cases. They both outperforms the SRS based estimates in accuracy, and the COA based estimate always yields the most accurate estimation. When the COA does not exist for a given value of $d$, we provide the novel idea of augmenting the coalition with additional null players so that the new number of players allows the construction of COA. This allows us to accommodate arbitrary number of players. 


\section*{Supplementary Material}\label{supp}

The supplementary material includes the algorithm for estimating the Shapley value in sensitivity analysis, all the proofs, and the simulation results for the voting game, sensitivity analysis and local model explanation.

\section*{Acknowledgements}
The authors would like to thank Valdis Krebs and Herbert Hamers for providing the data of the network of the WTC 9/11 attack. This work was partially supported by the National Natural Science Foundation of China (Grant Nos. 11871288, 12131001 and 12226343), Natural Science Foundation of Tianjin (19JCZDJC31100), Natural Science Foundation DMS-1830864, and National Ten Thousand Talents Program of China. The first two authors contributed equally to this work.


\end{document}